\documentclass[10pt,twocolumn,letterpaper]{article}

\usepackage{wacv}
\usepackage{times}
\usepackage{epsfig}
\usepackage{graphicx}
\usepackage{amsmath}
\usepackage{amssymb}

\usepackage{color}
\usepackage{siunitx}
\usepackage{multirow}
\usepackage{xcolor}
\usepackage{arydshln}

\newcommand{\extnode}{\emph{Leaf Event Node}}
\newcommand{\extnodes}{\emph{Leaf Event Nodes}}

\newcommand{\intnode}{\emph{Branch Event Node}}
\newcommand{\intnodes}{\emph{Branch Event Nodes}}

\newcommand{\set}{}

\newcommand{\veo}{\emph{Vi\-su\-al Event On\-tol\-o\-gy~\mbox{(VisE-O)}}}

\newcommand{\vecd}{\emph{Vi\-su\-al Event Clas\-si\-fi\-ca\-tion Da\-ta\-set~\mbox{(VisE-D)}}}

\def\wacvPaperID{578} 

\wacvfinalcopy 

\ifwacvfinal
\def\assignedStartPage{1} 
\fi

\ifwacvfinal
\usepackage[breaklinks=true,colorlinks,bookmarks=false]{hyperref}
\else
\usepackage[pagebackref=true,breaklinks=true,colorlinks,bookmarks=false]{hyperref}
\fi

\ifwacvfinal
\else
\fi

\begin{document}

\title{Ontology-driven Event Type Classification in Images}


\author{Eric Müller-Budack$^1$, Matthias Springstein$^1$, Sherzod Hakimov$^1$, Kevin Mrutzek$^2$, Ralph Ewerth$^{1,2}$ \\
$^1$TIB - Leibniz Information Centre for Science and Technology, Hannover, Germany \\
$^2$Leibniz University Hannover, L3S Research Center, Hannover, Germany \\
{\tt\small \{eric.mueller, matthias.springstein, sherzod.hakimov, ralph.ewerth\}@tib.eu} \\
}

\maketitle
\ifwacvfinal\thispagestyle{empty}\fi
\ifwacvfinal\pagestyle{empty}\fi

\begin{abstract}
%
Event classification can add valuable information for semantic search and the increasingly important topic of fact validation in news. So far, only few approaches address image classification for newsworthy event types such as natural disasters, sports events, or elections. Previous work distinguishes only between a limited number of event types and relies on rather small datasets for training.
In this paper, we present a novel ontology-driven approach for the classification of event types in images. We leverage a large number of real-world news events to pursue two objectives: First, we create an ontology based on Wikidata comprising the majority of event types. Second, we introduce a novel large-scale dataset that was 
acquired through Web crawling. 
Several baselines are proposed including an ontology-driven learning approach that aims to exploit structured information of a knowledge graph to learn relevant event relations using deep neural networks. 
Experimental results on existing as well as novel benchmark datasets demonstrate the superiority of the proposed ontology-driven approach.

\end{abstract}
\section{Introduction}
\label{sec:intro}
%
%
Digital media and social media platforms such as \emph{Twitter} have become a popular resource to provide news and information. To handle the sheer amount of daily published articles in the Web, automated solutions to understand the multimedia content are required. The computer vision community has focused on many visual classification tasks such as object recognition~\cite{DBLP:conf/cvpr/HeZRS16, DBLP:conf/iccv/HowardPALSCWCTC19, DBLP:conf/cvpr/HuangLMW17, DBLP:conf/nips/KrizhevskySH12, DBLP:conf/cvpr/ZophVSL18}, place~(scene) classification~\cite{DBLP:journals/pami/ZhouLKO018}, or geolocation estimation~\cite{DBLP:conf/eccv/Muller-BudackPE18, DBLP:conf/eccv/SeoWSH18, DBLP:conf/iccv/VoJH17, DBLP:conf/eccv/WeyandKP16} to enable semantic search or retrieval in archives and news collections. But news typically focus on events with a high significance for a target audience. Thus, event classification in images is an important task for various applications. 
Multimedia approaches~\cite{DBLP:conf/mm/JaiswalSAN17, DBLP:conf/mir/Muller-BudackTD20, DBLP:conf/mm/SabirA0N18} 
have exploited visual descriptors to quantify image-text relations that can help to understand the overall multimodal message and sentiment or might even indicate misinformation, i.e., \emph{Fake News}. 

Despite its clear potential, so far only few approaches~\cite{DBLP:conf/wacv/AhsanSHE17, DBLP:conf/iccv/BossardGV13, DBLP:conf/cvpr/JainSL08, DBLP:conf/iccv/LiF07, DBLP:conf/cvpr/XiongZLT15} were proposed 
for the classification of real-world event types. 
%
%
Datasets for event classification mostly cover only specific event categories, e.g., social~\cite{DBLP:conf/mmsys/AhmadCBN16, DBLP:conf/wacv/AhsanSHE17, DBLP:conf/mediaeval/ReuterPPMKCVG13}, sports~\cite{DBLP:conf/iccv/LiF07}, or cultural events~\cite{DBLP:conf/iccvw/EscaleraFPBGEMS15}. To the best of our knowledge, the \emph{Web Image Dataset for Event Recognition~(WIDER)}~\cite{DBLP:conf/cvpr/XiongZLT15} is the largest corpus with \num{50574} images that considers a variety of event types~(\num{61}). Nonetheless, many types that are important for news, like \emph{epidemics} or \emph{natural disasters}, are missing. 
Due to the absence of large-scale datasets, 
related work has focused on ensemble approaches~\cite{DBLP:conf/icip/AhmadMCBMN17, DBLP:journals/tomccap/AhmadMCMN18, DBLP:journals/ijcv/WangWQG18} typically based on pre-trained models for object and place~(scene) classification 
and the integration of descriptors from local image regions~\cite{DBLP:journals/spic/AhmadCN18, goebel2019deep, guo2020graph, DBLP:conf/cvpr/XiongZLT15} to learn rich features for event classification.   
%
%
We believe that one of the main challenges is to define a complete lexicon of important event categories.
For this purpose, Ahsan~\etal~\cite{DBLP:conf/wacv/AhsanSHE17} suggest to mine \emph{Wikipedia} and gathered \num{150} generic social events. However, the experiments were only conducted on \emph{WIDER} as well as on two datasets, which cover eight social event types and a selection of \num{21}~real-world events.
Progress in the field of Semantic Web has shown that it is possible to define a knowledge graph for newsworthy events~\cite{DBLP:conf/esws/GottschalkD18, DBLP:journals/semweb/GottschalkD19} but has not been leveraged by computer vision approaches yet. Particularly the relations between events extracted from a knowledge base such as \emph{Wikidata}~\cite{DBLP:journals/cacm/VrandecicK14} provide valuable information that can be utilized to train powerful models for event classification.
%


%
%
%
\begin{figure*}
\begin{center}
\includegraphics[width=1.0\linewidth]{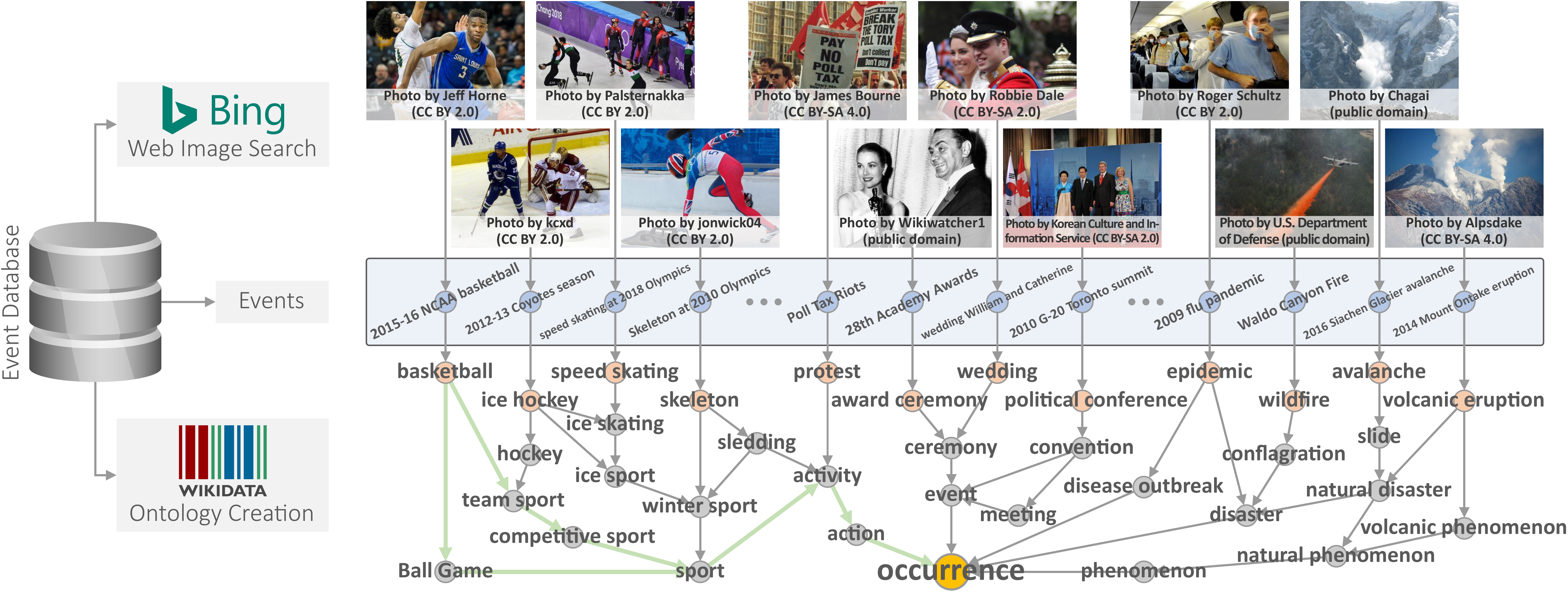}
\end{center}
\caption{Exemplary subset of the \emph{Ontology}~(complete version is provided on our \textit{GitHub} page\textsuperscript{\ref{foot:github}}) and images of the proposed \emph{Visual Event Classification Dataset~(VECD)}. \extnodes{}~(orange) and \intnodes{}~(gray) are extracted based on relations~(e.g., \emph{"subclass of"}) to a set of \emph{Events}~(blue) using the \emph{Wikidata} knowledge base. The nodes connected by the green path define the \emph{Subgraph} of \emph{basketball} to the \emph{Root Node}~(yellow). The combination~(union) of all \emph{Subgraphs} defines the \emph{Ontology}. Definitions are according to Section~\ref{sec:definitions}.} 
\label{fig:overview}
\end{figure*}
%
%
%

In this paper, 
we introduce a novel ontology along with a dataset that enable us to develop a novel ontology-driven deep learning approach for event classification. 
Our \textbf{primary contributions} can be summarized as follows:
(1)~Based on a set of real-world events from \emph{\mbox{\mbox{EventKG}}}~\cite{DBLP:conf/esws/GottschalkD18, DBLP:journals/semweb/GottschalkD19}, we propose a \veo{} containing \num{409} nodes describing \num{148}~unique event types such as different kinds of sports, disasters, and social events with high news potential that can be created with little supervision. It covers the largest number of event types for image classification to date. 
(2)~In order to train deep learning models, we have gathered a large-scale dataset, called \vecd{}, of \num{570540} images crawled automatically from the Web. It contains \num{531080} training and \num{28543} validation images as well as two test sets with \num{2779} manual annotated and \num{8138} \emph{Wikimedia} images. Figure~\ref{fig:overview} depicts some example images.
(3)~We provide several baselines including an ontology-driven deep learning approach that integrates the relations of event types extracted from structured information in the ontology to understand the fundamental differences of event types in different domains such as \emph{sports}, \emph{crimes}, or \emph{natural disasters}. 
Experimental results on several benchmark datasets demonstrate the feasibility of the proposed approach.
Dataset and source code are publicly available\footnote{
    Our project is available on the \textit{EventKG} website and on \textit{GitHub}: 
    
    \textit{EventKG}: \url{http://eventkg.l3s.uni-hannover.de/VisE} 
    
    \textit{GitHub}: \url{https://github.com/TIBHannover/VisE}
}\label{foot:github}

The remainder of this paper is organized as follows. In Section~\ref{chp:rw} we review related work. The ontology and dataset for newsworthy event types is presented in Section~\ref{chp:data_and_ont}. In Section~\ref{chp:cls} we propose an ontology-driven deep learning approach for event classification.
Experimental results for several benchmarks are presented in Section~\ref{chp:exp}. Section~\ref{chp:conc} summarizes the paper and outlines areas of future work.
\section{Related Work}
\label{chp:rw}
%
Since there are different definitions of an event, approaches for event classification are diverse and range from specific actions in videos~\cite{DBLP:conf/cvpr/TranWTRLP18, DBLP:conf/cvpr/XuYH15} over the classification of more personal events in photo collections~\cite{DBLP:journals/jvcir/BachaAB16, DBLP:conf/iccv/BossardGV13, DBLP:journals/tmm/WuHW15} to the classification of social, cultural, and sport events in photos~\cite{guo2020graph, DBLP:conf/iccv/LiF07, DBLP:journals/ijcv/WangWQG18, DBLP:conf/cvpr/XiongZLT15}. In the sequel, we mainly focus on works and datasets for the recognition of events and event types in images with potential news character.

Early approaches for event classification have used handcrafted features such as \emph{SIFT}~(Scale-Invariant Feature Transform) to classify events in particular domains like sports~\cite{DBLP:conf/cvpr/JainSL08, DBLP:conf/iccv/LiF07}. As one of the first deep learning approaches
Xiong~\etal~\cite{DBLP:conf/cvpr/XiongZLT15} trained a multi-layer framework that leverages two convolutional neural networks to incorporate the visual appearance of the whole image as well as interactions among humans and objects. Similarly, several approaches integrated local information from image patches or regions extracted by object detection frameworks~\cite{DBLP:journals/spic/AhmadCN18, goebel2019deep, guo2020graph} to learn rich features for event classification. In this respect, Guo~\etal~\cite{guo2020graph} proposed a graph convolutional neural network to leverage relations between objects. Another kind of approaches applies ensemble models and feature combination~\cite{DBLP:conf/icip/AhmadMCBMN17, DBLP:journals/tomccap/AhmadMCMN18, DBLP:journals/ijcv/WangWQG18} to exploit the capabilities of deep learning models trained for different computer vision tasks, most typically for object recognition and scene classification. In the absence of a large-scale dataset for many event types, Ahsan~\etal~\cite{DBLP:conf/wacv/AhsanSHE17} suggest to train classifiers based on images crawled for a set of \num{150}~social event concepts mined from \emph{Wikipedia}, while Wang~\etal~\cite{DBLP:journals/ijcv/WangWQG18} apply transfer learning to object and scene representations to learn compact representations for event recognition with few training images.
For a more detailed review of deep learning techniques for event classification, we refer to Ahmad and Conci's survey~\cite{DBLP:journals/tomccap/AhmadC19}.

There are many datasets and also challenges such as the \emph{Media\-Eval Social Event Detection Task}~\cite{DBLP:conf/mediaeval/ReuterPPMKCVG13} and \emph{ChaLearn Looking at People}~\cite{DBLP:conf/iccvw/EscaleraFPBGEMS15} for event classification. But they mostly cover specific domains such as social events~\cite{DBLP:conf/mmsys/AhmadCBN16,DBLP:conf/mediaeval/ReuterPPMKCVG13}, cultural events~\cite{DBLP:conf/iccvw/EscaleraFPBGEMS15}, or sports~\cite{DBLP:conf/iccv/LiF07}. In addition, the datasets are either too small~\cite{DBLP:conf/iccv/LiF07} to train deep learning models or contain very few event classes~\cite{DBLP:conf/mmsys/AhmadCBN16}. Other proposals have introduced datasets and approaches to detect concrete real-world news events~\cite{DBLP:conf/wacv/AhsanSHE17, DBLP:conf/iccvw/EscaleraFPBGEMS15, goebel2019deep}, but only distinguish between a small predefined selection.
To the best of our knowledge, \emph{WIDER~(Web Image Dataset for Event Recognition})~\cite{DBLP:conf/cvpr/XiongZLT15} is the most complete dataset in terms of the number of event categories that can be leveraged by deep learning approaches. It contains \num{50574} images for \num{61} event types. But many important event types for news such as \emph{epidemics} or \emph{natural disasters} are missing. 
%
\section{Ontology and Dataset}
\label{chp:data_and_ont}
In contrast to prior work, this section presents an ontology and dataset for event classification that covers a larger number of event types with news character across all domains such as \emph{sports}, \emph{crimes}, and \emph{natural disasters}.
%
%
Based on definitions for terms and notations~(Section~\ref{sec:definitions}), we suggest an approach that leverages events identified by \emph{\mbox{EventKG}}~\cite{DBLP:conf/esws/GottschalkD18, DBLP:journals/semweb/GottschalkD19} to automatically retrieve an ontology that can be refined with little supervision~(Section~\ref{sec:ont}). Images for event types in the resulting \veo{} are crawled from the Web to create the \vecd{} according to Section~\ref{sec:dataset}.
\subsection{Definitions and Notations}
\label{sec:definitions}
In this section, we introduce definitions and notations that are used in the remainder of the paper. Figure~\ref{fig:overview} contains supplementary visualizations to clarify the definitions.

\textbf{Event:} 
As in the \emph{\mbox{EventKG}}~\cite{DBLP:conf/esws/GottschalkD18}, we define a set $\set{E}$ of contemporary and historical events of global importance~(e.g., \emph{2011 NBA Finals} in Figure~\ref{fig:overview}) in this paper. 

\textbf{Ontology, Root Node, Event Node, and Relation:} 
The \emph{Ontology} is a directed graph composed by a set of \emph{Event Nodes}~$\set{N}$ and their corresponding \emph{Relations}~$\set{R}$ as edges. 
\emph{Relations}~$\set{R}$ are knowledge base specific properties such as \emph{"subclass of"} in \emph{Wikidata} that describe the interrelations of \emph{Event Nodes}~$\set{N}$.
All parent nodes~$n \in \set{N}$ that connect a specific \emph{Event}~$e \in \set{E}$ to the \emph{Root Node} are denoted as \emph{Event Nodes}.
The \emph{Root Node}~$n_R \in \set{N}$~(e.g., \emph{occurrence} in Figure~\ref{fig:overview}) matches the overall definition of an \emph{Event} and represents a parent node that is shared by all \emph{Events}.

\textbf{Leaf and Branch Event Node:} 
The \extnodes{} $\set{N}_L \subset \set{N}$ such as \emph{basketball} are the most detailed \emph{Event Nodes} without children in the \emph{Ontology}. They group \emph{Events} of the same type, e.g., \emph{2011 NBA Finals $\xrightarrow{}$ basketball}~(Figure~\ref{fig:overview}). 
\emph{Event Nodes}, e.g., \emph{ball game} with at least one child node are referred to as \intnodes{}~$\set{N}_B \subset \set{N}$.

\textbf{Subgraph:} A \emph{Subgraph}~$\set{S}_L$ is a set of all \emph{Event Nodes} $\set{S}_L = \{n_L, \dots, n_R\} \subset \set{N}$ that relate to a specified \extnode{}~$n_L \in \set{N}_L$ while traversing to the \emph{Root Node}~$n_R$.

\subsection{VisE-O: Visual Event Ontology}
\label{sec:ont}
%
%
\subsubsection{Knowledge Base and Root Node Selection} 
\begin{figure*}
	\begin{center}
    \includegraphics[width=0.872\linewidth]{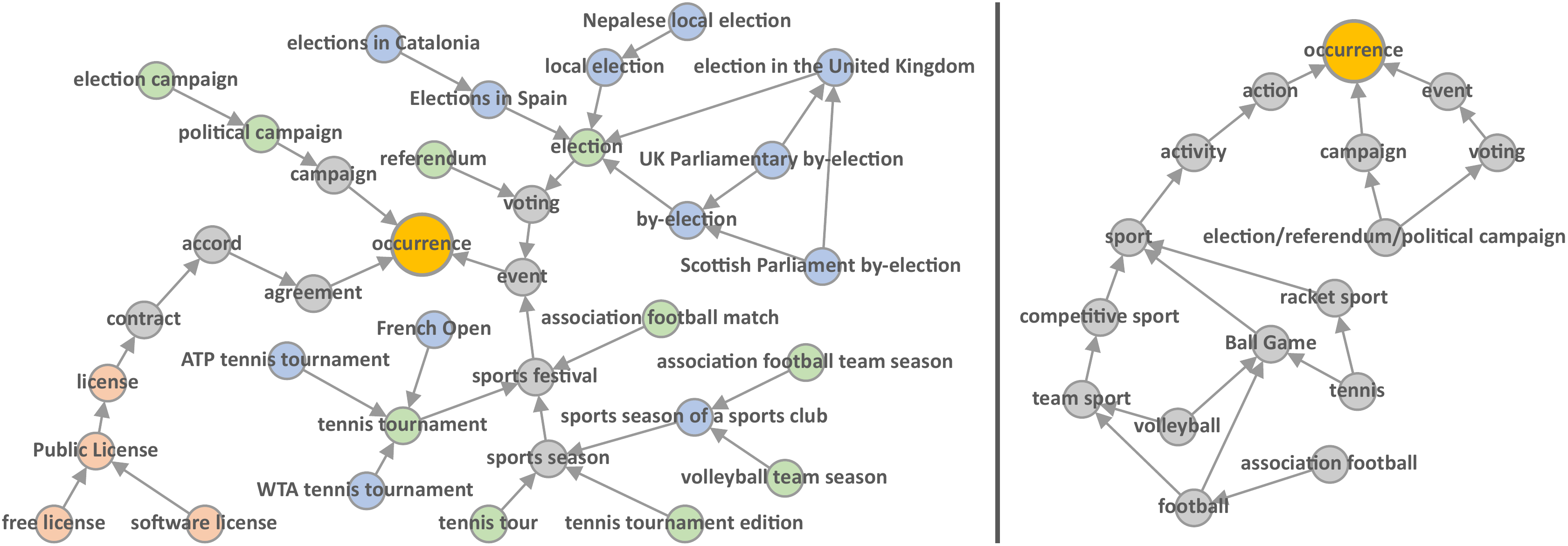}
    \end{center}
	\caption{Exemplary subset of the initial \emph{Ontology} after the extraction of all relations from \emph{Wikidata}~(left) and respective final \emph{Ontology} after applying the proposed approaches for event class disambiguation and refinement~(right). Blue \emph{Event Nodes} might be too fine-granular. Green nodes are semantically and visually similar to other \emph{Event Nodes} in the \emph{Ontology}. Orange nodes do not represent an \emph{Event} according to the definition in Section~\ref{sec:definitions}. Best viewed in color. Different versions of the ontologies can be explored on our \textit{GitHub} page\textsuperscript{\ref{foot:github}}.} 
	\label{fig:ontology}
\end{figure*}

Several knowledge bases such as \emph{DBpedia}~\cite{DBLP:conf/semweb/AuerBKLCI07}, \emph{YAGO}~\cite{DBLP:conf/www/SuchanekKW07}, or \emph{Wikidata}~\cite{DBLP:journals/cacm/VrandecicK14} are available. We investigated them in terms of event granularity and correctness. 
At this time, the whole \emph{DBpedia} ontology contains less than \num{1000}~classes. Thus, the granularity of potential event types is very coarse and for instance some types of natural disasters 
are either assigned to wrong~(\emph{Tsunami} $\xrightarrow{}$ \emph{television show})~\cite{IAtsunami} or generic classes~\mbox{(\emph{Earthquake} $\xrightarrow{}$ \emph{thing})}~\cite{IAearthquake}. As mentioned by 
Gottschalk and Demidova~\cite{DBLP:conf/esws/GottschalkD18}, \emph{YAGO} also contains noisy event categories. 
On the contrary, \emph{Wikidata} offers fine-granular event types and relations, as shown in Figure~\ref{fig:ontology}, and is therefore used as knowledge base in this work. We have selected \emph{occurrence~(Q1190554)} as the \emph{Root Node} of the \emph{Ontology} since it matches our definition of an \emph{Event}. 
%
%
%
\subsubsection{Initial Event Ontology}
\label{sec:init_ont}
In this paper, a \emph{bottom-up approach} is applied to automatically create an event ontology. Based on a large set of \mbox{$|\set{E}| = \num{550994}$} real-world events from \emph{\mbox{EventKG}}~\cite{DBLP:conf/esws/GottschalkD18, DBLP:journals/semweb/GottschalkD19}, we recursively obtain all parent \emph{Event Nodes} from \emph{Wikidata}. 
For \emph{Event Nodes} only relations of the type \emph{"subclass of"~(P279)} are considered since they already describe specific categories. For \emph{Events} we additionally allow the properties \emph{"instance of"~(P31)} and \emph{"part of"~(P361)} as possible relations to increase the coverage, because some events like 
\emph{2018 FIFA World Cup Group~A} are not a \emph{"subclass of"} an \emph{Event Node} but \emph{"part of"} a superordinate event, in this case \emph{2018 FIFA World Cup}. Finally, we remove all \emph{Event Nodes} that are not connected to the \emph{Root Node}. As illustrated in
Figure~\ref{fig:overview}, the resulting \emph{Subgraphs} 
define the \emph{Ontology}.

%
However, we identified several problems in the initial \emph{Ontology} as illustrated in Figure~\ref{fig:ontology}. 
(1)~There are differences in the granularity and some of the fine-grained \extnodes{}, e.g., \emph{ATP tennis tournament} or \emph{Nepalese local election}, might be hard to recognize; (2)~In particular, sports-centric \extnodes{} such as \emph{association football match} and \emph{association football team season} are ambiguous; 
(3)~Some \emph{Event Nodes}, e.g., \emph{software license} do not represent an \emph{Event} according to the definition in Section~\ref{sec:definitions}.
%
\subsubsection{Event Class Disambiguation}
As pointed out in the previous section, most \extnodes{} related to sports are visually ambiguous since they represent the same type of sport. The \emph{Wikidata} knowledge base distinguishes between \emph{sports seasons}, \emph{sports competitions}, etc. Although this structure might make sense for some applications, 
we aim to combine \emph{Event Nodes} that relate to the same sports type. 
Unfortunately, this is not possible with the initial \emph{Ontology} that relies on \emph{Relations} of the type \emph{"subclass of"}. 
As illustrated in Figure~\ref{fig:ontology}~(green nodes), \emph{Event Nodes} of different sports domains~(e.g., \emph{volleyball team season} and \emph{association football team season}) relate to a particular type of competition~(in this case~\emph{team season}) before they relate to another \emph{Event Node} of the same sports type~(\emph{association football match}). In order to solve this issue, value(s) for the \emph{Wikidata} property \emph{"sport"~(P641)}~(if available) for each \emph{Event} and \emph{Event Node} were extracted and used as \emph{Relation}. As a result, sports events were combined according to their sports category rather than the type of the competition as shown in Figure~\ref{fig:ontology}~(right). 
%
%
%
In addition, we delete all~\emph{Event Nodes} that are a parent of less than a minimum number of~$|\set{E}|_{min}=10$~\emph{Events} to reduce the granularity of the resulting \extnodes{}.

%
These strategies lead to an \emph{Ontology} that is more appropriate for computer vision tasks. However, it can still contain irrelevant \emph{Event Nodes}. Furthermore, scheduled events such as \emph{elections} or \emph{sports festivals} occur more frequently than unexpected or rare events such as \emph{epidemics} or \emph{natural disasters}. Therefore, \extnodes{} that represent scheduled event types more likely fulfill the filtering criteria~$|\set{E}|_{min}$ and are consequently very fine-grained~(e.g., elections in different countries) making them hard to distinguish. Thus, we decided to manually refine the \emph{Ontology}. 

\textbf{Clarification:} Please note, that related work also manually select the final set of classes~\cite{DBLP:conf/mmsys/AhmadCBN16, DBLP:conf/wacv/AhsanSHE17, DBLP:conf/iccvw/EscaleraFPBGEMS15, DBLP:conf/iccv/LiF07, DBLP:conf/cvpr/XiongZLT15}, 
whereas the proposed disambiguated \emph{Ontology} can be already used for many applications and provides a hierarchical overview to explore and select event classes more efficiently.  
%
\subsubsection{Event Ontology Refinement}
%
%
Two co-authors were asked to manually refine the \emph{Ontology} to create a challenging yet useful and fair \emph{Ontology} for image classification. To pursue this goal, the \emph{Ontology} was refined according to two criteria: (1)~reject \emph{Event Nodes} that do not match the \emph{Event} definition in Section~\ref{sec:definitions} and (2)~select the most suitable \extnodes{} to prevent ambiguities. 
For example, \emph{election} was chosen as a representative \extnode{} since its children contain different types of elections~(e.g., \emph{by-election}) and elections in different countries~(e.g., \emph{elections in Spain}) that might be too hard to distinguish. As we can use the hierarchical information to automatically assign the children to the selected \extnodes{} and simultaneously remove all resulting \intnodes{} as candidates, only around \num{500}~annotations were necessary to label all~(\num{2288}) \emph{Event Nodes}. 
Finally, we manually merged \num{30}~\extnodes{} such as 
\emph{award} and \emph{award ceremony} that are semantically similar but could not be fused using the \emph{Ontology}. 
\begin{table}
\begin{center}
\renewcommand{\b}{\textbf}
\renewcommand*{\arraystretch}{1.1}
\footnotesize
\setlength\tabcolsep{2pt}
\begin{tabular}{l | c c c c c | c c c c}
    \multirow{2}{*}{\b{Ontology}}   & \multicolumn{5}{c|}{\b{Ontology Statistics}} & \multicolumn{4}{c}{\b{Dataset Statistics}}\\
                    & $|\set{E}|$      & $|\hat{\set{E}}|$ & $|\set{N}|$      & $|\set{N}_L|$    & $|\set{R}|$      & $|\set{I}_T|$      & $|\set{I}_V|$     & $|\set{I}_B|$ & $|\set{I}_W|$\\ 
    \hline
    Initial         & \num{527}k & \num{236}k  & \num{6114} & \num{3578} & \num{7545} & ---          & ---         & ---     & --- \\
    Disamb.         & \num{530}k & \num{164}k  & \num{2288} & \num{1081} & \num{3144} & ---          & ---         & ---     & --- \\
    Refined         & \num{530}k & \num{447}k  & \num{409}  & \num{148}  & \num{635}  & \num{531}k   & \num{29}k & \num{2779} & \num{8138} \\
\end{tabular}
\end{center}
\caption{Number of \emph{Event Nodes}~$|\set{N}|$, \extnodes{}~$|\set{N}_L|$, \emph{Relations}~$|\set{R}|$, and images~$|\set{I}|$ for training~(T), validation~(V) and test~(B - \emph{VisE-Bing}, W - \emph{VisE-Wiki}). $|\set{E}|$~is the number of \emph{Events} that relate to any \emph{Event Node} in the \emph{Ontology}, and $|\hat{\set{E}}|$ the number of \emph{Events} that can be linked unambiguously to a \extnode{}.} 
\label{tab:ont_data}
\end{table}




%

The statistics for all variants of the \emph{Ontology} are shown in Table~\ref{tab:ont_data} and reveal that the refined \emph{Ontology} is able to link the most \emph{Events} to \extnodes{}. In the preliminary \emph{Ontologies}, many \emph{Events} are children of \intnodes{} and it is not possible to use them to query example images for \extnode{} as explained in the next section. The \emph{Ontologies} are provided on our \textit{GitHub} page\textsuperscript{\ref{foot:github}}.
\subsection{VisE-D: Visual Event Classification Dataset}
\label{sec:dataset}
%
\textbf{Data Collection:}
%
To create a large-scale dataset for the proposed \emph{Ontology} we defined different queries to crawl representative images from \emph{Bing}. A maximum of \num{1000}~images~(\num{500} without restrictions and another \num{500} uploaded within the last year) using the names of the \extnodes{} were crawled. In addition, the names of popular \emph{Events} related to a \extnode{} that happened after~1900 were used as queries to increase the number of images and reduce ambiguities~(e.g., \emph{Skeleton at the 2018 Winter Olympics} for \emph{Skeleton} in Figure~\ref{fig:overview}). In this regard, a sampling strategy~(details are provided on our \textit{GitHub} page\textsuperscript{\ref{foot:github}}) was applied to set the number of images downloaded for an \emph{Event} based on its popularity~(number of \emph{Wikipedia} page views) and date to prevent spam in the search results.

\textbf{Ground-truth Labels:} We provide two ground-truth vectors for each image based on the search query. (1)~The \textbf{\emph{Leaf Node Vector}}~$\mathbf{y}_L \in \{0, 1\}^{|\set{N}_L|}$ indicates which of the $|\set{N}_L|=148$~\extnodes{} are related to the image, and serves for classification tasks without using \emph{Ontology} information. Note that~$\mathbf{y}_L$ is multi-hot encoded as a queried \emph{Event}~(e.g, \emph{SpaceX Lunar Tourism Mission} $\rightarrow$ \emph{spaceflight} and \emph{expedition}) can relate to multiple \extnodes{}. (2)~The multi-hot encoded \textbf{\emph{Subgraph Vector}}~$\mathbf{y}_S \in \{0, 1\}^{|\set{N}|}$ denotes which of the $|\set{N}|=409$~\emph{Event Nodes}~(\emph{Leaf} and \emph{Branch}) are in the \emph{Subgraphs} of all related \extnodes{} and allows to learn from \emph{Ontology} information.

\textbf{Splits:}
We were able to download about \num{588000} images, which are divided into three splits for training~(90\%), validation~(5\%), and test~(5\%). For the test set we only use images from \emph{Events} that relate to exactly one \extnode{}. Test images that are a duplicate~(using the image hash) of a training or validation image are removed.  
%

\textbf{VisE-Bing Test Set:}
Two co-authors verified whether or not a test image depicts the respective \extnode{}. Each co-author annotated a maximum of ten valid images for each \extnode{} to prevent bias. They received different sets to increase the number of images. We obtained \num{2779} verified test images, with \num{20} images for most~(\num{109}) of the \num{148}~\extnodes{}. The dataset statistics are reported in Table~\ref{tab:ont_data} and in the supplemental material on \textit{GitHub}\textsuperscript{\ref{foot:github}}.
%

\textbf{VisE-Wiki Test Set:}
To create another larger test set, we downloaded all \emph{Wikimedia} images for each \extnode{} and its child \emph{Events} using the \emph{Commons category~(P373)} linked in \emph{Wikidata}. Despite \emph{Wikimedia} is a trusted source, we noticed some 
less relevant images for news, e.g., historic drawings or scans. We applied a k-nearest-neighbor classifier based on the embeddings of a \emph{ResNet-50}~\cite{DBLP:conf/cvpr/HeZRS16} trained on \emph{ImageNet}~\cite{DBLP:conf/cvpr/DengDSLL009}. For each test image in \emph{VisE-Bing}, we selected the $k=100/|I_a^{n}|$~nearest images, where $|\set{I}_a^{n}|$ is the number of annotated images of the \extnode{} \mbox{$n \in \set{N}_L$} in \emph{VisE-Bing}. The test set comprises \num{8138}~images for 146 of 148 classes~(statistics available on \textit{GitHub}\textsuperscript{\ref{foot:github}}). 
%
\section{Event Classification}
\label{chp:cls}
In this section, we propose a baseline classification approach~(Section~\ref{sec:cls_approach}) and more advanced strategies as well as weighting schemes to integrate event type relations from the \emph{Ontology} in the network training~(Section~\ref{sec:ont_approach}). 
%
%
\subsection{Classification Approach}
\label{sec:cls_approach}
%
As shown in Table~\ref{tab:ont_data} the refined \emph{Ontology} contains $|\set{N}_L| = 148$ \extnodes{}. As a baseline classifier, we train a convolutional neural network that predicts \extnodes{} without using ontology information. The \emph{Leaf Node Vector}~$\mathbf{y}_L = (y^{1}_L, \dots, y^{|\set{N}_L|}_L)$ from Section~\ref{sec:dataset} is used as target for optimization. During training the cross-entropy loss~$L_{c}$ based on the sigmoid activations~$\mathbf{\hat{y}}_L$ of the last fully-connected layer is optimized: 
\begin{equation}
\label{eq:cls_loss}
    L_{c} = -\sum\nolimits_{i = 1}^{|\set{N}_L|} y^{i}_L \cdot \log{\hat{y}^{i}_L}
\end{equation}
\subsection{Integration of Ontology Information}
\label{sec:ont_approach}
In order to integrate information from the proposed \emph{Ontology} in Section~\ref{sec:ont}, we use the multi-hot encoded \emph{Subgraph Vector}~$\mathbf{y}_{S} = (y^{1}_S, \dots, y^{|\set{N}|}_S)$ introduced in Section~\ref{sec:dataset} that includes the relations to all~$|\set{N}|=409$ \emph{Event Nodes} as a target.  
We consider two different loss functions. 
As for the classification approach, we apply the cross-entropy loss on the sigmoid activations~$\mathbf{\hat{y}}_{S}$ of last fully-connected layer to define an ontology-driven loss function: 
\begin{equation}
\label{eq:ont_loss}
    L^{cel}_{o} = -\sum\nolimits_{i = 1}^{|\set{N}|} y_{S}^{i} \cdot \log{\hat{y}_{S}^{i}} 
\end{equation}
As an alternative, we minimize the cosine distance of the predicted~$\mathbf{\hat{y}}_{S}$ and the ground truth~$\mathbf{y}_{S}$ \emph{Subgraph Vectors}: 
\begin{equation}
\label{eq:cos_loss}
    L^{cos}_{o} = 1 - \frac{\mathbf{y}_S \cdot \mathbf{\hat{y}}_S}{\left\|\mathbf{y}_S\right\|_2 \left\|\mathbf{\hat{y}}_S\right\|_2}
\end{equation}

However, the granularity and the number of \emph{Event Nodes} within the \emph{Subgraphs} of \extnodes{} varies for different domains such as \emph{sports}, \emph{elections}, or \emph{natural disasters}. As a consequence, the loss might be difficult to optimize. In addition, \intnodes{} such as \emph{action} or \emph{process} represent general concepts that are shared by many \extnodes{}. Some \intnodes{} are also redundant since they do not include more \extnodes{} as their corresponding children. 
\subsubsection{Redundancy Removal}
\label{sec:redundancy}
We delete every \intnode{} that relates to the same set of \extnodes{} compared to its child nodes in the \emph{Ontology}. These nodes are redundant since they do not include any new relationship information.
%
As a result, we are able to reduce the size of the \emph{Subgraph Vector}~$\mathbf{y}_S \in \{0, 1\}^{|\set{N}|}$ from~$|\set{N}|=\num{409}$ to $|\set{N}_{RR}|=\num{245}$.
\subsubsection{Node Weighting}
\label{sec:node_weight}
To encourage the neural network to focus on \extnodes{} and more informative \intnodes{} in the \emph{Ontology}, we investigated two weighting schemes. 
Based on \emph{one} of the schemes, each entry in the ground-truth~$\mathbf{y}_S$ and predicted~$\mathbf{\hat{y}}_S$ \emph{Subgraph Vectors} is multiplied with its corresponding weight before the loss according to Equation~\eqref{eq:ont_loss}~or~\eqref{eq:cos_loss} is calculated.

We propose a \textbf{Distance Weight}~$\gamma^{n}$ based on the distance of an \emph{Event Node}~$n \in \set{N}$ to all connected \extnodes{} in the \emph{Ontology}. First, the length~$l^n$ of the shortest path including self loops~(a node is always in its own path $l^n>0$) to each connected \extnode{} is determined. The average length~$\overline{l^n}$ of these paths is used to calculate the weight: 
\begin{equation}
\label{eq:node_dist_weight}
    \gamma^{n} = \frac{1}{2^{\left(\overline{l^n} - 1\right)}} \;.
\end{equation}

This weighting scheme encourages the network to learn from \emph{Event Nodes} that are close to the \extnodes{}. They describe more detailed event types which are harder to distinguish. Please note, that the average length $\overline{l^n}$ changes if the redundancy removal~(Section~\ref{sec:redundancy}) is applied.

Similarly, we calculate a \textbf{Degree of Centrality Weight}~$\omega_n$ for each \emph{Event Node}~$n \in \set{N}$ based on the number~$c^n$ of \extnodes{} connected to an \emph{Event Node}~$n$ and the total number of \extnodes{}~$|\set{N}_L| = 148$: 
\begin{equation}
\label{eq:node_weight}
    \omega^{n} = 1 - \frac{c^n - 1}{|\set{N}_L|} \;. 
\end{equation}

According to Equation~\eqref{eq:node_weight} the weights of all \extnodes{} are set to $\omega^{n} = 1, \forall n \in \set{N}_L$~(denoted as $\omega_L$), while, for instance, the \emph{Root Node}~$n_R$ is weighted with \mbox{$\omega^{n_R} \approx 0$} because it is connected to all \extnodes{}. 
Thus, the network should focus on learning unique event types such as \emph{tsunami} or \emph{carnival} rather than coarse superclasses that relate to many \extnodes{}. 
While the maximum weight of \intnodes{} using the \emph{Distance Weights} is~$0.5$ and defined by the nodes closest to the \extnodes{}~($\overline{l^n} = 2$), their corresponding \emph{Degree of Centrality Weight} can be close to $\omega_L$. To put more emphasis on \extnodes{}, we set their weights to $\omega_{L} > 1$. We set these weights to $\omega_{L} = 6$ as discussed in detail in Section~\ref{sec:ablation_study}.
\subsubsection{Inference}
\label{sec:infer}
The classification approach predicts a \emph{Leaf Node Vector}~$\mathbf{\hat{y}}_L$ that contains the probabilities of the $|\set{N}_L| = 148$~\extnodes{} that can be directly used for event classification. 
On the other hand, the ontology-driven network outputs a \emph{Subgraph Vector}~$\mathbf{\hat{y}}_S$ with probabilities for all $|\set{N}| = \num{409}$ or $|\set{N}_{RR}|=\num{245}$~(with redundancy removal)~\emph{Event Nodes} in the \emph{Ontology}. There are several options to retrieve a \emph{Leaf Node Vector}~$\mathbf{\hat{y}}_L$ for classification using~$\mathbf{\hat{y}}_S$. 

(1)~We retrieve the probabilities~$\mathbf{\hat{y}}^o_L$ that are part of the \emph{Subgraph Vector}~$\mathbf{\hat{y}}_S$ predicted by the ontology-driven approach.
(2)~Similar to Equation~\eqref{eq:cos_loss}, the cosine similarity of the predicted \emph{Subgraph Vector}~$\mathbf{\hat{y}}_S$ to the multi-hot encoded \emph{Subgraph Vector}~$\mathbf{y}^n_S$ of each \extnode{}~$n \in \set{N}_L$ is measured to leverage the probabilities of \intnodes{}. Note that the ground truth and predicted \emph{Subgraph Vectors} are first multiplied with the used weights during network training. As a result, we obtain $|\set{N}_L|=148$~similarities that are stored as~$\mathbf{\hat{y}}^{cos}_L \in \mathbb{R}^{|\set{N}_L|}$.

The elementwise product $\mathbf{\hat{y}}_L = \mathbf{\hat{y}}^o_L \odot \mathbf{\hat{y}}^{cos}_L$ is used as prediction for the ontology approach, we found that this combination worked best in most cases. Results using the individual probabilities can be found on our \textit{GitHub} page\textsuperscript{\ref{foot:github}}. 

%
%
\section{Experimental Setup and Results}
\label{chp:exp}
In this section, the utilized network architecture and parameters~(Section~\ref{sec:params}), evaluation metrics~(Section~\ref{sec:metrics}) as well as experimental results~(Section~\ref{sec:results}) are presented. 
%
%
%
\subsection{Network Parameters}
\label{sec:params}
In our experiments, we have used a \emph{ResNet-50}~\cite{DBLP:conf/cvpr/HeZRS16} as the basic architecture for the proposed approaches. The networks were optimized using \emph{Stochastic Gradient Descent~(SGD)} with \emph{Nesterov momentum} term~\cite{DBLP:conf/icml/SutskeverMDH13}, weight decay of~\num{1e-5}, and a batch size of \num{128}~images. 
To speed-up the training, the initial learning rate of \num{0.01} is increased to \num{0.1} using a linear ramp up in the first \num{10000} iterations. Subsequently, a cosine learning rate annealing~\cite{DBLP:conf/iclr/LoshchilovH17} is applied to lower the learning rate to zero after a total of \num{100000} iterations. Finally, the model that achieves the lowest loss on the validation set is used for the experiments. 
\subsection{Evaluation Metrics}
\label{sec:metrics}
%
%
%
We report the top-1, top-3, and top-5 accuracy using the top-k predictions in the \emph{Leaf Node Vector}~$\mathbf{\hat{y}}_L$~(Section~\ref{sec:infer}). 
%
%
But the accuracy does not reflect the similarity of the predicted to the ground-truth \extnode{} with respect to the \emph{Ontology} information.
For this reason, we create a multi-hot encoded \emph{Subgraph Vector}~$\mathbf{y}_{\hat{S}} \in \{0, 1\}^{|\set{N}|}$ representing the whole \emph{Subgraph}~$\hat{\set{S}}$ of the predicted~(top-1) \extnode{}~$\hat{n}$. 
Note, that the full \emph{Subgraph Vector} with dimension $|\set{N}|=409$ is created to generate comparable results for models trained with and without redundancy removal. 
We propose to measure the cosine similarity~(\emph{CS}; similar to Equation~\eqref{eq:cos_loss}) and \emph{Jaccard Similarity Coefficient}~(\emph{JSC}; Equation~\eqref{eq:jaccard})  between $\mathbf{y}_{\hat{S}} $ and the ground-truth \emph{Subgraph Vector}~$\mathbf{y}_{S}$ of the test image to quantify the similarity based on all $|\set{N}|=409$ \emph{Event Nodes}:
\begin{equation}
    \label{eq:jaccard}
    \textnormal{\emph{JSC}} = \frac{\left\|\mathbf{y}_S \odot \mathbf{y}_{\hat{S}}\right\|_1}{\left\|\mathbf{y}_S\right\|_1 \cdot \left\|\mathbf{y}_{\hat{S}}\right\|_1 \cdot \left\|\mathbf{y}_S \odot \mathbf{y}_{\hat{S}}\right\|_1}
\end{equation}

\subsection{Experimental Results}
\label{sec:results}
In this section, the results of our proposed approaches are presented. First, we compare a variety of ontology-driven approaches to the classification baseline in Section~\ref{sec:ablation_study}. A detailed analysis of the results for specific event types is presented in Section~\ref{sec:indepth_results}. Finally, we evaluate the proposed approaches on other benchmarks~(Section~\ref{sec:sota_comparison}).

%
%
\renewcommand{\cm}{\checkmark}
\renewcommand{\b}{\textbf}

\begin{table}
\begin{center}    
\renewcommand*{\arraystretch}{1.1}
\footnotesize
\setlength\tabcolsep{1.7pt}
\begin{tabular}{l | c | c | c | c c c | c c }
& \multirow{2}{*}{\b{Loss}} & \multirow{2}{*}{\b{WS}} & \multirow{2}{*}{\b{RR}} & \multicolumn{3}{c|}{\b{Accuracy}} & \multirow{2}{*}{\b{\emph{JSC}}} & \multirow{2}{*}{\b{\emph{CS}}} \\ 
                        & & & & \emph{\b{Top1}} & \emph{\b{Top3}} & \emph{\b{Top5}} & \\
    \hline
    $C$                         & $L_{c}$       & &                                 & 77.4 & 89.8 & 93.6 & 84.7 & 87.7 \\
    \hline
    $O^{cel}$                   & $L_{o}$               &                           &       & 67.5 & 83.3 & 88.5 & 81.1 & 85.4 \\
    \hdashline
    $O^{cel}_{\omega}$          & $L_{o}^{cel}$         & $\omega$, $\omega_{L}=1$  &       & 68.1 & 83.7 & 88.9 & 81.1 & 85.3 \\
    $O^{cel}_{6\omega}$         & $L_{o}^{cel}$         & $\omega$, $\omega_{L}=6$  &       & 79.8 & 91.0 & 94.0 & 86.6 & 89.2 \\
    $O^{cel}_{6\omega}$+$RR$    & $L_{o}^{cel}$         & $\omega$, $\omega_{L}=6$  & \cm   & 81.7 & 91.5 & \b{94.5} & \b{87.9} & 90.3 \\
    \hdashline
    $O^{cel}_{\gamma}$          & $L_{o}^{cel}$         & $\gamma$                  &       & 66.6 & 83.5 & 89.1 & 78.3 & 82.8 \\
    $O^{cel}_{\gamma}$+$RR$     & $L_{o}^{cel}$         & $\gamma$                  & \cm   & 73.2 & 86.8 & 91.3 & 82.6 & 86.2 \\
    \hline
    $O^{cos}$                   & $L_{o}^{cos}$         &                           &       & 67.6 & 77.8 & 81.8 & 82.6 & 86.7 \\
    \hdashline
    $O^{cos}_{\omega}$          & $L_{o}^{cos}$         & $\omega$, $\omega_{L}=1$  &       & 72.7 & 84.1 & 87.2 & 84.5 & 87.9 \\
    $O^{cos}_{6\omega}$         & $L_{o}^{cos}$         & $\omega$, $\omega_{L}=6$  &       & 80.2 & 90.6 & 93.4 & 86.3 & 88.9 \\
    $O^{cos}_{6\omega}$+$RR$    & $L_{o}^{cos}$         & $\omega$, $\omega_{L}=6$  & \cm   & 80.8 & 90.1 & 93.1 & 86.9 & 89.4 \\
    \hdashline
    $O^{cos}_{\gamma}$          & $L_{o}^{cos}$         & $\gamma$                  &       & 81.1 & 90.2 & 93.1 & 87.1 & 89.7 \\
    $O^{cos}_{\gamma}$+$RR$     & $L_{o}^{cos}$         & $\gamma$                  & \cm   & 80.7 & 90.3 & 93.1 & 86.9 & 89.5 \\
    \hline
    $CO^{cel}_{6\omega}$+$RR$   & $L_{c}+L_{o}^{cel}$   & $\omega$, $\omega_{L}=6$  & \cm   & 81.5 & \b{91.8} & 94.3 & 87.5 & 90.0 \\
    $CO^{cos}_{\gamma}$         & $L_{c}+L_{o}^{cos}$   & $\gamma$                  &       & \b{81.9} & 90.8 & 93.2 & \b{87.9} & \b{90.4} \\
\end{tabular}
\end{center}
\caption{Results on \emph{VisE-Bing} using different loss functions, weighting schemes~(WS), and ontology redundancy removal~(RR).}
\label{tab:ablation_VECD}
\end{table}

\subsubsection{Ablation Study}
\label{sec:ablation_study}
The results of the proposed approaches on \emph{VisE-Bing} are presented in Table~\ref{tab:ablation_VECD}.
The performances of the ontology-driven approaches are significantly worse without applying any weighting scheme, because the correct prediction of the majority of \emph{Event Nodes} in a \emph{Subgraph} is already sufficient to achieve low loss signals. However, the ontology-driven approaches greatly benefit from the weighting schemes and clearly outperform the classification baseline. As discussed in Section~\ref{sec:node_weight}, a higher weight~$\omega_{L}$ for \extnodes{} needs to be assigned using the \emph{Degree of Centrality Weights} to balance the impact of \emph{Branch} and \emph{Leaf Event Nodes} on the overall loss. Thus, we increased the weight to~$\omega_{L}=6$ as it approximately corresponds to the average number of \intnodes{} in all $|\set{N}_L|=148$~\emph{Subgraphs}.

Both loss functions~$L_{o}^{cel}$ and ~$L_{o}^{cos}$ achieve similar results in their best setups. Models trained with $L_o^{cos}$ work well with both weighting schemes, while models optimized with $L_o^{cel}$ are better with the \emph{Degree of Centrality Weight}. We believe they are more tailored towards single-label classification tasks and benefit from the higher weights~$\omega_{L}=6$ of \extnodes{}.
%
We were able to achieve slightly better results combining both loss functions, since it puts more emphasis on the prediction of \extnodes{} while still considering ontology information. 
 
The best results with respect to \emph{top-1 accuracy}, \emph{JSC}, and \emph{CS} were achieved by combining the classification and ontology-driven cosine loss term with \emph{Distance Weights}. The cosine loss is in general more stable when training with and without redundancy, which could indicate that it is more robust to changes in depth and size of the \emph{Ontology}. Furthermore it works well with the \emph{Distance Weight} which does not require an extra weight for \extnodes{}.

%
\subsubsection{Performance for Individual Event Types}
\label{sec:indepth_results}
\begin{figure}
	\begin{center}
    \includegraphics[width=0.95\linewidth,clip]{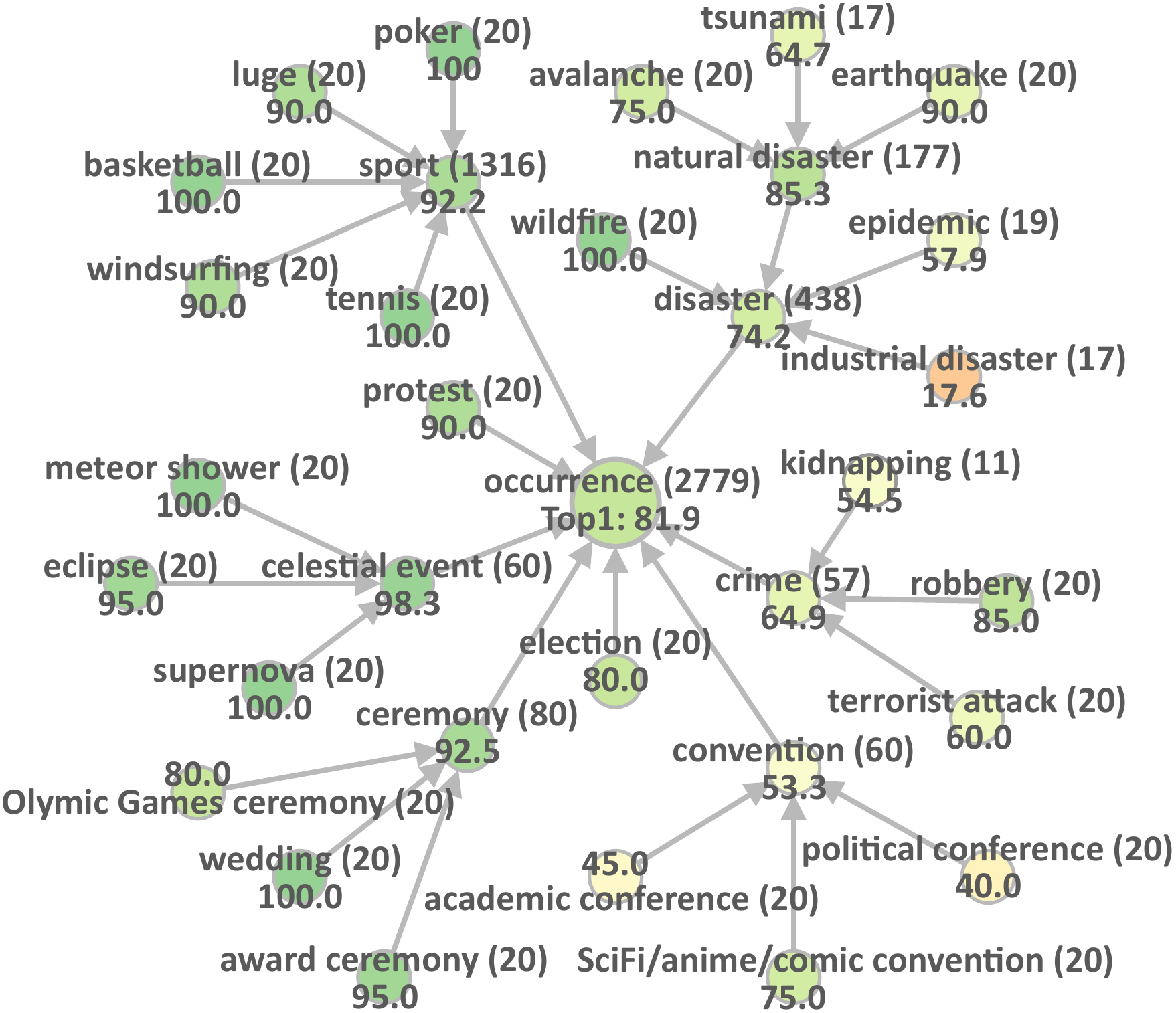}
    \end{center}
	\caption{\textit{Top-1 accuracy} and number of images~(in brackets) for a selection of \emph{Event Nodes} on \emph{VisE-Bing} using the $CO^{cos}_{\gamma}$ approach. The results correspond to the mean \textit{top-1 accuracy} of all~(also those that are not shown) related \extnodes{}. The \emph{Ontology} is simplified for better comprehensibility.} 
	\label{fig:confmat}
\end{figure}
\begin{figure}
	\begin{center}
    \includegraphics[width=1.00\linewidth]{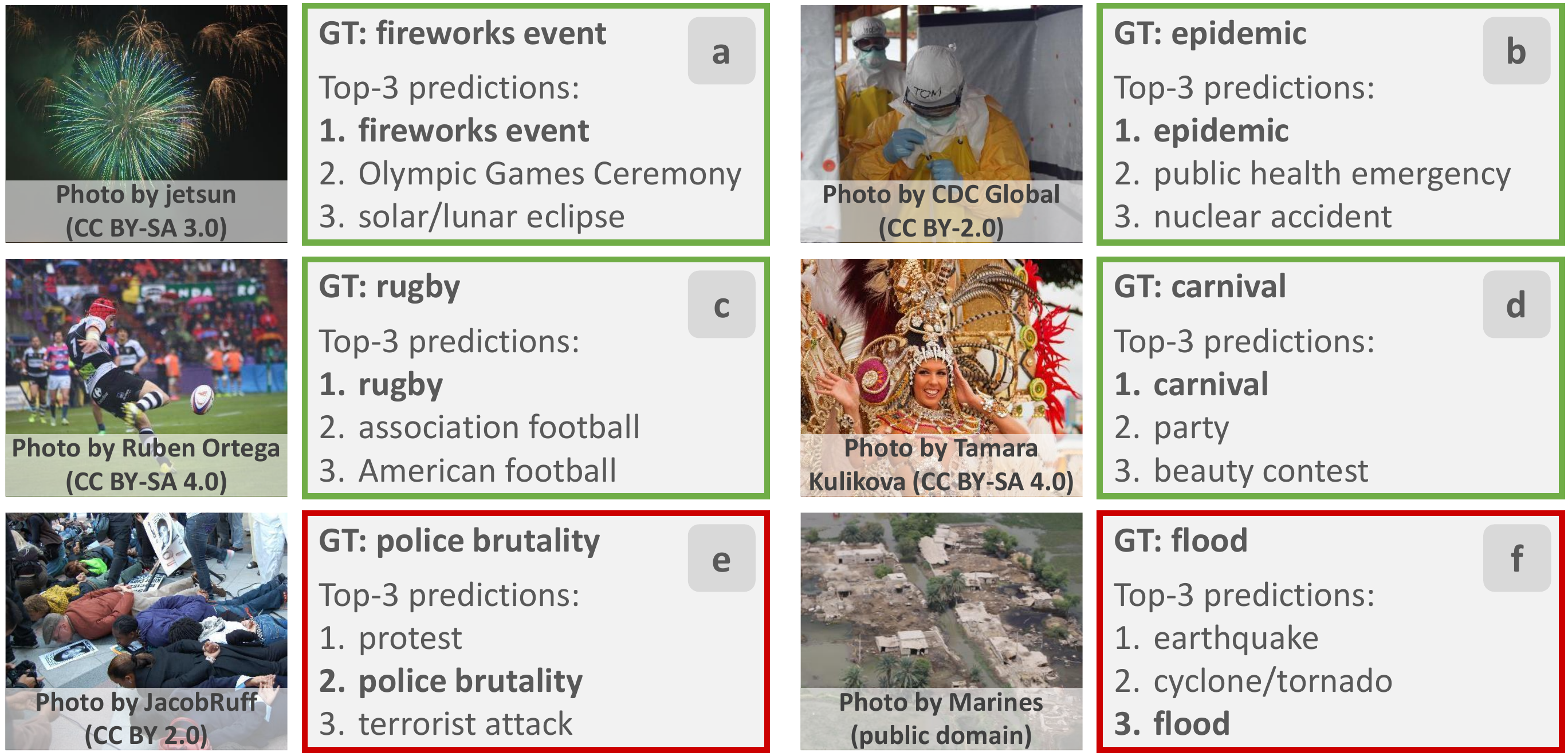}
	\end{center}
    \caption{Correctly~(green) and incorrectly~(red) classified examples of the $CO^{cos}_{\gamma}$ network model from \emph{VisE-Wiki}.}
	\label{fig:qual_results}
\end{figure}
The \textit{top-1 accuracy} for a selection of \emph{Event Nodes} and qualitative results of the $CO_{\gamma}^{cos}$ model are provided in Figure~\ref{fig:confmat}~and~\ref{fig:qual_results}. The proposed approach achieves good results for a majority of event types. 
Misclassification can be typically explained by the visual similarity of the respective events. For example, images for \emph{tornado}, \emph{tsunami}, and \emph{earthquake} are often captured after the actual event and the consequences of these natural disasters can be visually similar as illustrated in Figure~\ref{fig:overview} and \ref{fig:qual_results}f. 
It also turned out, that classes such as \emph{protest}, \emph{earthquake}, and \emph{explosion} are predicted very frequently, because they depict visual concepts that are also part of other events. For instance, images of the event types \emph{police brutality}, \emph{vehicle fire}, and \emph{economic crisis} are frequently classified as \emph{protest} since they depict typical scenes of riots or demonstrations~(Figure~\ref{fig:qual_results}e). The best results were achieved for sports-centric event types, which is not surprising as they are usually unambiguous.
In general, the performance for scheduled event types such as \emph{election} and \emph{sport} is better compared to unexpected or rare events. We assume the main reason is that journalists usually broadcast live coverage of scheduled events, while photos of crimes~(e.g., \emph{robbery}, \emph{terrorist attack}) and \emph{natural disasters} are rare and mostly captured by amateurs. 
\subsubsection{Comparisons on other Benchmarks}
\label{sec:sota_comparison}
We considered several benchmarks including the novel \emph{VisE-Wiki}~(Section~\ref{sec:dataset}) test dataset as well as \emph{WIDER}~\cite{DBLP:conf/cvpr/XiongZLT15}, \emph{SocEID}~\cite{DBLP:conf/wacv/AhsanSHE17}, and \emph{RED}~\cite{DBLP:conf/wacv/AhsanSHE17}. These benchmarks have different characteristics, which allows us to evaluate the ontology-driven approach in various setups. 
%
\emph{WIDER} comprises \num{50574} Web images for \num{61}~event types. 
The \emph{Social Event Image Data\-set~(SocEID)} consists of circa \num{37000} images but contains only eight social event classes, while \emph{Rare Events Dataset~(RED)} is comparatively small and contains around \num{7000} images from \num{21} real-world events. We used the splits provided by the authors for \emph{WIDER}~\cite{DBLP:conf/cvpr/XiongZLT15} and \emph{SocEID}~\cite{DBLP:conf/wacv/AhsanSHE17}. For \emph{RED} we randomly used $70\%$ of the dataset for training and $30\%$ for testing, as suggested by Ahsan~\etal~\cite{DBLP:conf/wacv/AhsanSHE17}. The splits are provided\textsuperscript{\ref{foot:github}} to allow fair comparisons.

As \emph{WIDER}, \emph{SocEID}, and \emph{RED} do not provide an \emph{Ontology}, we have manually linked the classes~(e.g., \emph{soccer} to \emph{association football~(Q2736)} in \emph{WIDER}) to \emph{Wikidata} to define the set of \extnodes{}. Then, we created the \emph{Ontologies}~(provided on our \textit{GitHub} repository\textsuperscript{\ref{foot:github}}) according to Section~\ref{sec:init_ont}. 
The models are mostly trained with the parameters from Section~\ref{sec:params}. Due to the smaller dataset sizes the number of training iterations was reduced to \num{2500} for \emph{RED} and \num{10000} for \emph{SocEID} and \emph{WIDER}. Cosine learning rate annealing~\cite{DBLP:conf/iclr/LoshchilovH17} was applied from the beginning to lower the learning rate from \num{0.01} to zero after the specified amount of iterations. 
%
The results for our approach and other comparable solutions that use a single network and the whole image as input are presented in Table~\ref{tab:sota_comp}.
%

\renewcommand{\g}{\color[gray]{0.5}}
\renewcommand{\b}{\textbf}

\begin{table}
\label{tab:sota_comp}
\begin{center}
\renewcommand*{\arraystretch}{1.1}
\footnotesize
\setlength\tabcolsep{3pt}
\begin{tabular}{l | c c | c c | c c | c c }
    \multirow{3}{*}{\b{Approach}} & \multicolumn{2}{c|}{\b{VisE-Wiki}} & \multicolumn{2}{c|}{\b{WIDER~\cite{DBLP:conf/cvpr/XiongZLT15}}} & \multicolumn{2}{c|}{\b{SocEID~\cite{DBLP:conf/wacv/AhsanSHE17}}} & \multicolumn{2}{c}{\b{RED~\cite{DBLP:conf/wacv/AhsanSHE17}}} \\
    
    & \multicolumn{2}{c|}{\b{148 classes}} & \multicolumn{2}{c|}{\b{61 classes}} & \multicolumn{2}{c|}{\b{8 classes}} & \multicolumn{2}{c}{\b{21 classes}} \\
    
    & \emph{\b{Top1}} & \emph{\b{JSC}} & \emph{\b{Top1}} & \emph{\b{JSC}} & \emph{\b{Top1}} & \emph{\b{JSC}} & \emph{\b{Top1}} & \emph{\b{JSC}} \\
    \hline
    AlexNet~\cite{DBLP:conf/cvpr/XiongZLT15}            & --- & ---             & 38.5  & --- & --- & --- & --- & --- \\
    \hline
    AlexNet-fc7~\cite{{DBLP:conf/wacv/AhsanSHE17}}      & --- & ---             & \g{77.9} & --- & 86.4  & --- & \g{77.9} & --- \\
    WEBLY-fc7~\cite{{DBLP:conf/wacv/AhsanSHE17}}        & --- & ---             & \g{77.9} & --- & 83.7  & --- & \g{79.4} & --- \\
    Event conc.~\cite{{DBLP:conf/wacv/AhsanSHE17}}      & --- & ---             & \g{78.6} & --- & 85.4  & --- & \g{77.6} & --- \\
    \hline
    AlexNet~\cite{DBLP:journals/tomccap/AhmadMCMN18}  & --- & ---             & 41.9  & --- & --- & --- & --- & --- \\
    ResNet152~\cite{DBLP:journals/tomccap/AhmadMCMN18}  & --- & ---             & 48.0  & --- & --- & --- & --- & --- \\
    \hline
    \hline
    $C$                                                 & 61.7 & 72.7           & 45.6 & 56.9           & 91.2 & 92.7           & 76.1 & 82.1 \\
    \hdashline
    $CO^{cel}_{6\omega}$+$RR$                           & 63.4 & 73.9           & \b{51.0 }& \b{61.6}   & 91.4 & \b{92.9}       & 79.1 & 84.3 \\
    $CO^{cos}_{\gamma}$                                 & \b{63.5} & \b{74.1}   & 49.7 & 60.3           & \b{91.5} & \b{92.9}   & \b{80.9} & \b{85.4} \\
\end{tabular}
\end{center}
\caption{Results on different benchmarks. While our results are superior on \emph{SocEID} and \emph{RED}, Ahsan~\etal~\cite{DBLP:conf/wacv/AhsanSHE17} achieved better results~(77.9\%) on \emph{WIDER} using random splits~(gray, not provided on request) also compared to other baselines by training a SVM on \emph{AlexNet} embeddings, which is a similar approach for which Ahmad~\etal~\cite{DBLP:journals/tomccap/AhmadMCMN18} reported 41.9\%. Their results for \emph{WIDER} and \emph{RED} are nearly identical, although \emph{WIDER} contains more classes and is in general more challenging. We believe these results are not explainable and need to be verified in a reproducibility experiment.}

\end{table}

The ontology-driven approaches~($CO$) clearly outperform the classification baseline~($C$) on \emph{VisE-Wiki}, \emph{WIDER}, and \emph{RED}. As expected, the results on \emph{SocEID} just slightly improved, because less \emph{Ontology} information are provided due to the lower number of eight classes and thus \emph{Event Nodes}. 
Compared to the results for \emph{VisE-Bing}~(Table~\ref{tab:ablation_VECD}), the performances are worse on \emph{VisE-Wiki}, because the test set is not manually annotated and contains noisy or ambiguous imagery, particularly for rare event types such as \emph{city fire}. The same applies for \emph{WIDER}.
%
%
Superior performances are achieved in comparison to similar solutions. It is worth noting that the proposed ontology-driven approach can also be easily integrated in frameworks that utilize ensemble models~\cite{DBLP:conf/icip/AhmadMCBMN17,DBLP:journals/tomccap/AhmadMCMN18,DBLP:journals/ijcv/WangWQG18} or additional image regions~\cite{guo2020graph,DBLP:conf/cvpr/XiongZLT15}.
%
%
%
\section{Conclusions and Future Work}
\label{chp:conc}
%
In this paper, we have presented a novel ontology, dataset, and ontology-driven deep learning approach for the classification of newsworthy event types in images. A large number of events in conjunction with a knowledge base were leveraged to retrieve an ontology that covers many possible real-world event types. The corresponding large-scale dataset with \num{570540} images allowed us to train powerful deep learning models and is, to the best of our knowledge, the most complete and diverse public dataset for event classification to date.
We have proposed several baselines including an ontology-driven deep learning approach that exploits event relations to integrate structured information from a knowledge graph. 
The results on several benchmarks have shown that the integration of structured information from an ontology can improve event classification. 

In the future, we plan to further explore strategies to leverage ontology information such as graph convolutional networks. Other interesting research directions are the combination of several knowledge bases and the investigation of semi-supervised approaches to learn from noisy Web data. 

\section*{Acknowledgement}
This project has received funding from the European Union’s Horizon 2020 research and innovation programme under the Marie Skłodowska-Curie grant agreement no 812997~(project name: "CLEOPATRA"), and the German Research Foundation (DFG: Deutsche Forschungsgemeinschaft) under "VIVA - Visual Information Search in Video Archives"~(project number: 388420599).
{\small
\bibliographystyle{ieee_fullname}
\bibliography{egbib}
}

\end{document}